\documentclass{article}
\usepackage{geometry}
\usepackage[margin=1cm]{caption}
\usepackage{blindtext}
\usepackage[toc,page]{appendix}
\usepackage[utf8x]{inputenc}
\usepackage[T1]{fontenc}
\usepackage{url}
\usepackage{adjustbox}
\usepackage{multirow}
\usepackage{algorithm} 
\usepackage{algpseudocode}

\title{Evaluating KGR10 Polish word embeddings in the recognition of temporal expressions using BiLSTM-CRF.}

\author{Jan Kocoń, Michał Gawor\\{\small\rm Department of Computational Intelligence} \\ Faculty of Computer Science and Management \\ Wroclaw University of Technology \\ e-mail: {\it \{jan.kocon,michal.gawor\}@pwr.edu.pl}}

\begin{document}
\maketitle

\abstract{The article introduces a new set of Polish word embeddings, built using KGR10 corpus, which contains more than 4 billion words. These embeddings are evaluated in the problem of recognition of temporal expressions (timexes) for the Polish language.  We described the process of KGR10 corpus creation and a new approach to the recognition problem using Bidirectional Long-Short Term Memory (BiLSTM) network with additional CRF layer, where specific embeddings are essential. We presented experiments and conclusions drawn from them.}


\section{Introduction}
Recent studies in information extraction domain (but also in other natural language processing fields) show that deep learning models produce state-of-the-art results~\cite{young2018}. Deep architectures employ multiple layers to learn hierarchical representations of the input data. In the last few years, neural networks based on dense vector representations provided the best results in various NLP tasks, including named entities recognition~\cite{santos2015}, semantic role labelling~\cite{he2017}, question answering~\cite{yu2014} and multitask learning~\cite{collobert2008}. The core element of most deep learning solutions is the dense distributed semantic representation of words, often called \emph{word embeddings}. Distributional vectors follow the distributional hypothesis that words with a similar meaning tend to appear in similar contexts. Word embeddings capture the similarity between words and are often used as the first layer in deep learning models. Two of the most common and very efficient methods to produce word embeddings are \emph{Continuous Bag-of-Words} (CBOW) and \emph{Skip-gram} (SG), which produce distributed representations of words in a vector space, grouping them by similarity~\cite{mikolov2013a, mikolov2013}. With the progress of machine learning techniques, it is possible to train such models on much larger data sets, and these often outperform the simple ones. It is possible to use a set of text documents containing even billions of words as training data. Both architectures (CBOW and SG) describe how the neural network learns the vector word representations for each word. In CBOW architecture the task is \emph{predicting the word given its context} and in SG the task in \emph{predicting the context given the word}. 

Due to a significant increase of quality using deep learning methods together with word embeddings as the input layer for neural networks, many word vector sets have been created, using different corpora. The widest range of available word embeddings is available for English~\cite{kutuzov2017} and there were not so many options for less popular languages, e.g.\ Polish. There was a definite need within CLARIN-PL\footnote{\url{https://clarin-pl.eu/}} project and Sentimenti\footnote{\url{https://sentimenti.pl/}} to increase the quality of NLP methods for Polish which were utilising available Polish word vectors~\cite{piasecki2018, bojanowski2017, mykowiecka2017a, rogalski2016} but only FastText modification of Skip-gram~\cite{bojanowski2017} was able to produce vectors for unknown words, based on character n-grams. The observation was that even using a sophisticated deep neural structure, the result strongly depends on the initial distributional representation. There was a need to build a massive corpus of Polish and create high-quality word vectors from that corpus. This work describes how we extended KGR7 1G corpus to become KGR10 with 4 billion words. Next, we present the different variants of word embeddings produced using this corpus. In the article about the recognition of named entities for Polish from the previous year, these embeddings were used in one of the three voting models to obtain the best results and the final system PolDeepNer~\cite{Marcinczuk2018} took the second place in PolEval2018 Task 2~\cite{ogrodniczuk2018}. In this article, we evaluated KGR10 FastText word embeddings in recognition of timexes.

\section{Available word embeddings}
At the time we were testing word embeddings for different applications, there were 2 most popular sources of word vectors. The first one, called \textbf{IPIPAN}\footnote{\url{http://dsmodels.nlp.ipipan.waw.pl/}}, is the result of the project \emph{Compositional distributional semantic models for identification, discrimination and disambiguation of senses in Polish texts}, the process of creating word embeddings is described in article~\cite{mykowiecka2017a} and corpora used were National Corpus of Polish (NKJP)~\cite{przepiorkowski2012} and Wikipedia (Wiki). The second one, called \textbf{FASTTEXT}\footnote{\url{https://fasttext.cc/docs/en/crawl-vectors.html}}, is original FastText word embeddings set, created for 157 languages (including Polish). Authors used Wikipedia and Common Crawl\footnote{\url{http://commoncrawl.org/}} as the linguistic data source. Table~\ref{tab:corpora} shows the number of tokens in each corpus and the name of the institution which prepared it. There is also information about the public availability of the resource.

\begin{table}[ht]
	
	\center
	\begin{tabular}{|cllrrc|}
		\hline
		C\_ID & Corpus       & Prepared by & Tokens          &  Unique words      & Open \\ \hline
		C1 & Wikipedia    & FASTTEXT    & 386,874,622     & 1,298,250   & yes \\
		C2 & Common Crawl & FASTTEXT    & 21,859,939,298  & 10,209,556  & yes \\
		C3 & Wikipedia\_2    & IPIPAN      & $\sim$184,000,000   & $\sim$3,000,000 & yes \\
		C4 & NKJP         & IPIPAN      & $\sim$1,044,000,000 & $\sim$8,200,000 & no \\
		\hline
	\end{tabular}      
	\caption{Informations about corpora used to prepare embeddings by FASTTEXT and IPIPAN: corpus ID, number of tokens, number of unique words, the name of the institution and the availability of the resource.}
	
	\label{tab:corpora}
\end{table}

Table~\ref{tab:embeddings} presents the most commonly used word embeddings in CLARIN-PL before the creation of our embeddings. 

\begin{table}[ht]
	
	\center
	\begin{tabular}{|ccllrcl|}
		\hline
			E\_ID & S\_IDs & Name                        & Method & Dimension  & Binary  & Prepared by \\ \hline
			EE1           &  C1, C2    & cc.pl.300                         & CBOW   & 300  &  yes     & FASTTEXT \\
			EE2 			 &  C3, C4    & NWfa-1-s-n  & SG     & 100  &  no      & IPIPAN \\
			EE3           &  C3, C4    & NWfa-3-s-n  & SG     & 300  &  no      & IPIPAN \\
		\hline
	\end{tabular}      
	\caption{Available word embeddings (external, EE -- created outside Wroclaw University of Technology, G4.19 Group) with the information about embedding ID, linguistic sources used to create embedding, original embedding name, method of creation, vector dimension, format and the institution which prepared the resource. Original file names are: \mbox{cc.pl.300} -- \texttt{cc.pl.300.bin}, \mbox{NWfa-1-s-n} -- \texttt{
		nkjp+wiki-forms-all-100-skipg-ns.vec}, \mbox{NWfa-3-s-n} -- \texttt{nkjp+wiki-forms-all-300-skipg-ns.vec}}
	
	\label{tab:embeddings}
\end{table}

\section{Building a larger corpus}
KGR7 corpus (also called plWordNet Corpus 7.0, PLWNC~7.0)~\cite{kedzia2013,maziarz2016} was created at the Wroclaw University of Science and Technology by G4.19 Group. Due to the licences of documents in this corpus, this resource is not publicly available. Table~\ref{tab:kgr7} contains KGR7 subcorpora and statistics~\cite{Wendelberger2015}. One of the subcorpora in KGR7 is KIPI (the IPI PAN Corpus)~\cite{przepiorkowski2004}. KGR7 covers texts from a wide range of domains like: blogs, science, stenographic recordings, news, journalism, books and parliamentary transcripts. All texts come from the second half of the 20th century and represent the modern Polish language.

\begin{table}[ht]
	
	\center
	\begin{tabular}{|lr|}
		\hline
		Subcorpus name & Tokens \\ \hline
		1002 & 19,512,317 \\
		1003 & 	10,006,539 \\
		blogi & 9,613,618 \\
		interia & 611,402\\
		kipi & 255,516,328\\
		knigi joined & 1,010,676,150\\
		naukawe & 2,594,225\\
		ornitologia & 544,937\\
		plwiki20120428 & 275,578,635\\
		pogoda & 593,538\\
		poig biznes data sub0& 35,439,099\\
		poig biznes data sub1& 30,676,362\\
		polityka & 82,480,654\\
		prace & 12,665,419\\
		pryzmat & 2,183,403\\
		rzepa & 116,317,357\\
		sjp & 2,177,299\\
		wordpress & 439,304\\
		zwiazki & 820,991\\ \hline
		SUM & 1,868,447,577			\\
		\hline
	\end{tabular}      
	\caption{Names and the number of tokens in KGR7 subcorpora.}
	
	\label{tab:kgr7}
\end{table}  

\subsection{plWordNet Corpus 10.0 (KGR10)}
KGR10, also known as plWordNet Corpus 10.0 (PLWNC~10.0), is the result of the work on the toolchain to automatic acquisition and extraction of the website content, called CorpoGrabber\footnote{\url{http://hdl.handle.net/11321/403}}~\cite{kocon2014}. It is a pipeline of tools to get the most relevant content of the website, including all subsites (up to the user-defined depth). The proposed toolchain can be used to build a big Web corpus of text documents. It requires the list of the root websites as the input. Tools composing CorpoGrabber are adapted to Polish, but most subtasks are language independent. The whole process can be run in parallel on a single machine and includes the following tasks: download of the HTML subpages of each input page URL with HTTrack\footnote{\url{https://www.httrack.com}}, extraction of plain text from each subpage by removing boilerplate content (such as navigation links, headers, footers, advertisements from HTML pages)~\cite{pomikalek2011}, deduplication of plain text~\cite{pomikalek2011}, bad quality documents removal utilising Morphological Analysis Converter and Aggregator (MACA)~\cite{radziszewski2011}, documents tagging using Wrocław CRF Tagger (WCRFT)~\cite{radziszewski2013}. Last two steps are available only for Polish. 

In order to significantly expand the set of documents in KGR7, we utilised DMOZ (short for \emph{directory.mozilla.org}) -- a multilingual open content directory of World Wide Web links, also known as Open Directory Project (ODP). The website with directory was closed in 2017, but the database still can be found on the web. Polish part of this directory contains more than 30,000 links to Polish websites. We used these links as root URLs for CorpoGrabber, and we downloaded more than 7TB of HTML web pages.  After the extraction of text from HTML pages, deduplication of documents (including texts from KGR7) and removing bad quality documents (containing more than 30\% of words outside the Morfeusz~\cite{wolinski2014} dictionary) the result is KGR10 corpus, which contains 4,015,569,051 tokens and 18,084,712 unique words. Due to component licenses, KGR10 corpus is not publicly available.

\section{KGR10 word embeddings}
We created a new Polish word embeddings models using the KGR10 corpus. We built 16 models of word embeddings using the implementation of CBOW and Skip-gram methods in the FastText tool~\cite{bojanowski2017}. These models are available under an open license in the CLARIN-PL project DSpace repository\footnote{\url{https://clarin-pl.eu/dspace/handle/11321/606}}. The internal encoding solution based on embeddings of n-grams composing each word makes it possible to obtain FastText vector representations, also for words which were not processed during the creation of the model. A vector representation is associated with character n-gram and each word is represented as the sum of its n-gram vector representations. Previous solutions ignored the morphology of words and were assigning a distinct vector to each word. This is a limitation for languages with large vocabularies and many rare words, like Turkish, Finnish or Polish~\cite{bojanowski2017}. Authors observed that using word representations trained with subword information outperformed the plain Skip-gram model and the improvement was most significant for morphologically rich Slavic languages such as Czech (8\% reduction of perplexity over SG) and Russian (13\% reduction)~\cite{bojanowski2017}. We expected that word embeddings created that way for Polish should also provide such improvements. There were also previous attempts to build KGR10 word vectors with other methods (including FastText), and the results are presented in the article~\cite{piasecki2018}. We selected the best models from that article  -- with \emph{embedding ID} prefix \textbf{EP} (embeddings, previous) in Table~\ref{tab:embeddings_kgr10} -- to compare with new models, marked as \emph{embedding ID} prefix \textbf{EC} in Table~\ref{tab:embeddings_kgr10}). 

\begin{table}[ht]
	
	\center
	\begin{tabular}{|llcccc|}
		\hline
		E\_ID  & Name        &  Dim.  & Bin. & Meth. & App.  \\ \hline
		EP1    & skip\_gram\_v100m8\_nomwe.w2v.vec       & 100 & 0 & S & ns            \\
		EP2    & cbow\_v100m8\_nomwe.w2v.vec             & 100 & 0 & C & ns             \\
		EP3    & cbow\_v100m8\_hs\_nomwe.w2v.vec         & 100 & 0 & C & hs              \\
		EP4    & cbow\_v100m8\_hs.w2v.vec                & 100 & 0 & C & hs              \\
		EP5    & skip\_gram\_v100m8.w2v.vec              & 100 & 0 & S & ns              \\
		EP6    & cbow\_v100m8.w2v.vec                    & 100 & 0 & C & ns              \\
		EP7    & skip\_gram\_v300m8.w2v.vec              & 300 & 0 & S & ns              \\
		EP8    & cbow\_v300m8\_hs.w2v.vec                & 300 & 0 & C & hs              \\
		EP9    & cbow\_v300m8.w2v.vec                    & 300 & 0 & C & ns              \\
		EP10    & kgr10-plain-sg-300-mC50.bin            & 300 & 0 & S & ns              \\ \hline
		EC1    & kgr10.plain.skipgram.dim300.neg10.bin   & 300 & 1 & S & ns              \\
		EC2    &  kgr10.plain.skipgram.dim100.neg10.bin  & 100 & 1 & S & ns              \\
		EC3    &  kgr10.plain.cbow.dim100.neg10.bin  & 100 & 1 & C & ns              \\
		EC4    &  kgr10.plain.cbow.dim300.neg10.bin  & 300 & 1 & C & ns              \\
		\hline
	\end{tabular}      
	\caption{KGR10 word embeddings created at WUST, G4.19, with the information about embedding ID (EP -- previous, EC -- current), original embedding name, dimension, binary format, method of creation (\textbf{S}kipgram, \textbf{C}BOW), softmax approximation method (hs -- hierarchical softmax, ns -- negative sampling).}
	
	\label{tab:embeddings_kgr10}
\end{table}  

The word embeddings models used in PolDeepNer for recognition of timexes and named entities were EE1, . It was built on a plain KGR10. The dimension of word embedding is 300, the method of constructing vectors was Skip-gram~\cite{bojanowski2017}, and the number of negative samples for each positive example was 10.  

\section{Temporal expressions}
\label{sec:intro}
Temporal expressions (henceforth \emph{timexes}) tell us \emph{when} something happens, \emph{how long} something lasts, or \emph{how often} something occurs. The correct interpretation of a timex often involves knowing the context. Usually, a person is aware of their location in time, i.e., they know what day, month and year it is, and whether it is the beginning or the end of week or month. Therefore, they refer to specific dates, using incomplete expressions such as \emph{12 November}, \emph{Thursday}, \emph{the following week}, \emph{after three days}. The temporal context is often necessary to determine to which specific date and time timexes refer. These examples do not exhaust the complexity of the problem of recognising timexes. 

TimeML~\cite{Sauri2006} is a markup language for describing timexes that has been adapted to many languages. One of the best-known methods of recognition of timexes called \emph{HeidelTime}~\cite{Strotgen2013}, which uses the TIMEX3 annotation standard, currently supports 13 languages (with the use of hand-crafted resources). 
PLIMEX is a specification for the description of Polish timexes. It is based on TIMEX3 used in TimeML. Classes proposed in TimeML are adapted, namely: \emph{date}, \emph{time}, \emph{duration}, \emph{set}.

\section{Recognition of timexes}
There are many methods for recognising timexes that are widely used in natural language engineering. For English (but not exclusively), in approaches based on supervised learning, sequence labelling methods are often used, especially Conditional Random Fields~\cite{Lafferty2001}. A review of the methods in the article~\cite{Uzzaman2013} about the recognition of timexes for English and Spanish has shown a certain shift within the most popular solutions. As with the normalisation of timexes, the best results are still achieved with rule-based methods, many new solutions have been introduced in the area of recognition. The best systems listed in~\cite{Uzzaman2013}, called \emph{TIPSem}~\cite{Llorens2010a} and \emph{ClearTK}~\cite{Bethard2013}, use CRFs for recognition, so initially, we decided to apply the CRF-based approach for this task. The results were described in~\cite{Kocon2015, Kocon2015a}.

In recent years, solutions based on deep neural networks, using word representation in the form of word embeddings, created with the use of large linguistic corpus, have begun to dominate in the field of recognition of word expressions. The most popular solutions include bidirectional long short-term memory neural networks (henceforth Bi-LSTM), often in combination with conditional random fields, as presented in the paper~\cite{Habibi2017} dedicated to the recognition of proper names. For the Polish language, deep networks have also recently been used to recognise word expressions. In the issue of recognition of timexes, a bidirectional gated recurrent unit network (GRU) has been used~\cite{Mykowiecka2017,Mykowiecka2018}. GRU network is described in detail in the article~\cite{Chung2014}. In case of recognition of event descriptions using Bi-LSTM and Bi-GRU, where most of the Liner2 features were included in the input feature vector, better results were obtained~\cite{Kobylinski2018} than for the Liner2 method (but without taking into account domain dictionaries). In last year's publication on the issue of named entities recognition using BiLSTM+CRF (together with G4.19 Group\footnote{\url{http://nlp.pwr.edu.pl/}} members), we received a statistically significant improvement in the quality of recognition compared to a solution using CRF only. The solution has been called PolDeepNer\footnote{\url{https://github.com/CLARIN-PL/PolDeepNer}}~\cite{Marcinczuk2018}.

\section{Experiments and Results}
Experiments were carried out by the method proposed in~\cite{Uzzaman2013}. The first part is described as \emph{Task A}, the purpose of which is to identify the boundaries of timexes and assign them to one of the following classes: \emph{date, time, duration, set}.

  \begin{table}[h]
  	\center
  	\begin{tabular}{|lrr|}
  		\hline
  		\textbf{Data set} & \textbf{Documents} & \textbf{Part [\%]} \\ \hline
  		all & 1635 & 100 \\ 
  		train & 1227 & 50 \\ 
  		test & 408 & 25 \\ 
  		\hline
  	\end{tabular}      
  	\label{tab:data_size}
  	\caption{Evaluation data sets (source: KPWr).}
  \end{table}
  
We trained the final models using the \emph{train} set and we evaluated it using the \emph{test} set, which was the reproduction of analysis performed in articles~\cite{Kocon2016a, Kocon2017}. The division is presented in Table~\ref{tab:data_size}. We used BiLSTM+CRF classifier as in previous work~\cite{Marcinczuk2018}. We used \emph{precision, recall and F1} metrics from the classic NER task~\cite{Marcinczuk2018}, where \emph{true positive} system answer has the same boundaries \emph{and} type as annotation in gold data set. We evaluated all 17 word embeddings models using these metrics. The results are presented in Tables~\ref{tab:deepner_eval_precision},~\ref{tab:deepner_eval_recall}~and~\ref{tab:deepner_eval_fscore}.

We chose the best 3 results from each word embeddings group (EE, EP, EC) from Table~\ref{tab:deepner_eval_fscore} presenting F1-scores for all models. Then we evaluated these results using more detailed measures for timexes, presented in~\cite{Uzzaman2013}. The following measures were used to evaluate the quality of boundaries and class recognition, so-called strict match: \emph{strict precision (Str.P)}, \emph{strict recall (Str.R)} and \emph{strict F1-score (Str.F1)}. A \emph{relaxed match (Rel.P, Rel.R, Rel.F1)} evaluation has also been carried out to determine whether there is an overlap between the system entity and gold entity, e.g. \emph{[Sunday]} and \emph{[Sunday morning]}~\cite{Uzzaman2013}. If there was an overlap, a \emph{relaxed type F1-score (Type.F1)} was calculated~\cite{Uzzaman2013}. The results are presented in Table~\ref{tab:deepner_eval_timeml}.

\begin{table}[ht]
	\center
	\begin{tabular}{|lccccc|}
		\hline		
		\textbf{Embedding} &
		\textbf{Date} &
		\textbf{Duration}  &
		\textbf{Set} & 
		\textbf{Time}  &
		\textbf{Total}  \\ \hline
		EE1  & 93.87 & 81.37 & 83.87 & 79.39 & 90.33                         \\ 
		EE2  & 93.72 & 76.85 & 87.27 & 73.34 & 91.19                         \\ 
		EE3  & 95.14 & 83.25 & 86.00 & 77.16 & 91.34                         \\ \hline
		EP1  & 92.52 & 73.52 & 66.10 & 76.76 & 87.36                         \\ 
		EP2  & 91.53 & 72.85 & 75.00 & 68.52 & 85.93                         \\ 
		EP3  & 94.83 & 73.76 & 76.09 & 75.17 & 89.15                         \\ 
		EP4  & 92.75 & 70.09 & 76.19 & 60.12 & 89.15                         \\  
		EP5  & 93.83 & 77.29 & 83.64 & 68.94 & 88.54                         \\ 
		EP6  & 93.41 & 69.26 & 78.33 & 68.59 & 86.75                         \\ 
		EP7  & 92.57 & 73.42 & \textbf{90.00}& 73.91 & 88.04                 \\ 
		EP8  & 91.76 & 78.76 & 79.17 & 76.27 & 88.41                         \\ 
		EP9  & 92.50 & 75.86 & 86.84 & 73.28 & 88.37                         \\ 
		EP10 & 94.69 & 77.68 & 80.82 & 79.27 & 90.00                         \\ \hline
		EC1  & 96.00 & 81.70 & 85.29 & 76.80 & \textbf{91.37}                \\ 
		EC2  & 94.06 & 80.84 & 81.03 & \textbf{80.92}& 90.37                 \\ 
		EC3  & 92.82 & 76.50 & 73.85 & 68.50 & 86.78                         \\ 
		EC4  & \textbf{96.15}& \textbf{84.36}& 71.64 & 71.26 & 90.85 \\ 

		\hline
	\end{tabular}      
	\caption{Evaluation results (precision) for 17 word embeddings models for each TIMEX3 class (date, time, duration and set).}
	
	\label{tab:deepner_eval_precision}
\end{table}

\begin{table}[ht]
	\center
	\begin{tabular}{|lccccc|}
		\hline		
		\textbf{Embedding} &
		\textbf{Date} &
		\textbf{Duration}  &
		\textbf{Set} & 
		\textbf{Time}  &
		\textbf{Total}  \\ \hline
		EE1  & 94.44 & 76.50 & 65.00 & 75.72 & 88.53                                    \\ 
		EE2  & 93.47 & 76.50 & 60.00 & 73.99 & 87.09                                    \\ 
		EE3  & 93.12 & 77.88 & 53.75 & 72.25 & 86.85                                    \\ \hline
		EP1  & 61.62 & 74.19 & 48.75 & 63.01 & 84.04                                    \\ 
		EP2  & 90.48 & 74.19 & 48.75 & 64.16 & 83.35                                    \\
		EP3  & 88.89 & 75.12 & 43.75 & 63.01 & 81.98                                    \\
		EP4  & 90.21 & 75.58 & 60.00 & 60.12 & 81.98                                    \\
		EP5  & 89.86 & 73.73 & 57.50 & 64.16 & 83.29                                    \\
		EP6  & 91.27 & 73.73 & 57.75 & 61.85 & 84.10                                    \\
		EP7  & 92.24 & 45.12 & 45.00 & 58.96 & 83.98                                    \\
		EP8  & 90.30 & 70.05 & 45.50 & 52.02 & 81.30                                    \\
		EP9  & 90.30 & 70.97 & 41.25 & 55.49 & 81.48                                    \\
		EP10 & 94.36 & 83.41 & \textbf{73.75}& 75.14 & 89.78                            \\ \hline
		EC1  & \textbf{96.33}& \textbf{84.33}& 72.50 & 80.35 & \textbf{91.08}           \\
		EC2  & 94.89 & 79.72 & 58.75 & \textbf{80.92}& 89.53                            \\
		EC3  & 93.47 & 82.49 & 60.00 & 79.19 & 88.78                                    \\ 
		EC4  & 94.71 & 82.03 & 60.00 & 68.79 & 88.47                                    \\ 
		\hline
	\end{tabular}      
	\caption{Evaluation results (recall) for 17 word embeddings models for each TIMEX3 class (date, time, duration and set).}
	
	\label{tab:deepner_eval_recall}
\end{table}

\begin{table}[ht]
	\center
	\begin{tabular}{|lccccc|}
		\hline		
		\textbf{Embedding} &
		\textbf{Date} &
		\textbf{Duration}  &
		\textbf{Set} & 
		\textbf{Time}  &
		\textbf{Total}  \\ \hline
		EE1  & 94.15 & 78.86 & 73.24 & 77.51 & 89.42                           \\ 
		EE2  & 93.60 & 76.67 & 71.11 & 73.56 & 88.15                           \\ 
		EE3  & 94.12 & 80.48 & 66.15 & 74.63 & 89.04                           \\ \hline
		EP1  & 92.07 & \textbf{83.85}& 56.12 & 69.21 & 85.67                   \\
		EP2  & 91.00 & 73.52 & 59.09 & 66.27 & 84.62                           \\
		EP3  & 91.76 & 74.43 & 55.56 & 68.55 & 85.42                           \\
		EP4  & 91.46 & 72.73 & 67.13 & 60.12 & 84.29                           \\
		EP5  & 91.80 & 75.47 & 68.15 & 66.47 & 85.83                           \\
		EP6  & 92.33 & 71.43 & 67.14 & 65.05 & 85.41                           \\
		EP7  & 92.40 & 74.26 & 60.00 & 65.59 & 85.96                           \\ 
		EP8  & 91.02 & 74.15 & 59.37 & 61.86 & 84.70                           \\
		EP9  & 91.39 & 73.33 & 55.93 & 61.16 & 84.78                           \\
		EP10 & 94.52 & 80.44 & 77.12 & 77.15 & 89.89                           \\ \hline
		EC1  & \textbf{95.66}& 82.99 & \textbf{78.38}& 78.53 & \textbf{91.23}  \\
		EC2  & 94.47 & 80.28 & 68.12 & \textbf{80.92}& 89.95                   \\
		EC3  & 93.15 & 79.38 & 66.21 & 73.46 & 87.77                           \\ 
		EC4  & 95.42 & 83.18 & 65.31 & 70.00 & 89.64                           \\ 
		\hline
	\end{tabular}      
	\caption{Evaluation results (F1-score) for 17 word embeddings models for each TIMEX3 class (date, time, duration and set).}
	
	\label{tab:deepner_eval_fscore}
\end{table}  

\begin{table}[ht]
	\small
	\center
	\begin{tabular}{|lccccccc|}
		\hline		
		\textbf{Embedding} &
		\textbf{Str.P} &
		\textbf{Str.R}  &
		\textbf{Str.F1} & 
		\textbf{Rel.P}  &
		\textbf{Rel.R}  &
		\textbf{Rel.F1}  &
		\textbf{Type.F1}  \\ \hline
		EE1  & 91.31 & 88.40 & 89.83 & 96.07 & 93.02 & 94.52 & 92.24 \\ 
		EE2  & 90.56 & 90.34 & 90.45 & 95.38 & 95.14 & 95.26 & 92.88 \\ 
		EE3  & 91.15 & 90.52 & 90.84 & 95.79 & 95.14 & 95.46 & 93.09 \\ \hline
		EP5  & 91.13 & 82.67 & 86.69 & 96.56 & 87.59 & 91.86 & 89.44 \\ %
		EP7  & 89.41 & 84.73 & 87.00 & 95.26 & 90.27 & 92.70 & 90.08 \\ %
		EP10 & 90.52 & 90.52 & 90.52 & 95.45 & 95.45 & 95.45 & 92.52 \\ \hline %
		EC1  & 91.97 & \textbf{92.77}& \textbf{92.36}& 95.92 & \textbf{96.76}& \textbf{96.34}& \textbf{94.54}\\
		EC2  & 92.28 & 90.90 & 91.58 & \textbf{96.96}& 95.51 & 96.23 & 94.28 \\
		EC4  & \textbf{92.41}& 90.27 & 91.33 & 96.81 & 94.58 & 95.68 & 93.22 \\ 
		\hline
	\end{tabular}      
	\caption{Evaluation results for all TIMEX3 classes (total) for 9 word embeddings models (3 best models from each embeddings group: EE, EP, EC from Table~\ref{tab:deepner_eval_fscore}) using the following measures from~\cite{Uzzaman2013}: \emph{strict precision, strict recall, strict F1-score, relaxed precision, relaxed recall, relaxed F1-score, type F1-score}.}
	
	\label{tab:deepner_eval_timeml}
\end{table}

\section{Conclusions}
The analysis of results from Tables~\ref{tab:deepner_eval_precision},~\ref{tab:deepner_eval_recall}~and~\ref{tab:deepner_eval_fscore} show that 12 of 15 best results were obtained using new word embeddings. The evaluation results presented in Table~\ref{tab:deepner_eval_timeml} (the chosen best embeddings models from Table~\ref{tab:deepner_eval_fscore}) 
prove that the best group of word embeddings is EC. The highest \emph{type F1-score} was obtained for EC1 model, built using binary FastText Skip-gram method utilising subword information, with vector dimension equal to 300 and negative sampling equal to 10. The ability of the model to provide vector representation for the unknown words seems to be the most important. Also, previous models built using KGR10 (EP) are probably less accurate due to an incorrect tokenisation of the corpus. We used WCRFT tagger~\cite{radziszewski2013}, which utilises Toki~\cite{radziszewski2011} to tokenise the input text before the creation of the embeddings model. The comparison of EC1 with previous results obtained using only CRF~\cite{Kocon2017} show the significant improvement across all the tested metrics: 3.6pp increase in \emph{strict F1-score}, 1.36pp increase in \emph{relaxed precision}, 5.61pp increase in \emph{relaxed recall} and 3.51pp increase in \emph{relaxed F1-score}. 

\section*{Acknowledgements}
Work co-financed as part of the investment in the CLARIN-PL research infrastructure funded by the Polish Ministry of Science and Higher Education and in part by the National Centre for Research and Development, Poland, under grant no
POIR.01.01.01-00-0472/16.


\bibliographystyle{plain}
\bibliography{tfml}

\begin{thebibliography}{10}

\bibitem{Bethard2013}
Steven Bethard.
\newblock {ClearTK-TimeML: A minimalist approach to TempEval 2013}.
\newblock pages 10--14, June 2013.

\bibitem{bojanowski2017}
Piotr Bojanowski, Edouard Grave, Armand Joulin, and Tomas Mikolov.
\newblock Enriching word vectors with subword information.
\newblock {\em Transactions of the Association for Computational Linguistics},
  5:135--146, 2017.

\bibitem{Chung2014}
Junyoung Chung, Caglar Gulcehre, KyungHyun Cho, and Yoshua Bengio.
\newblock Empirical evaluation of gated recurrent neural networks on sequence
  modeling.
\newblock {\em arXiv preprint arXiv:1412.3555}, 2014.

\bibitem{collobert2008}
Ronan Collobert and Jason Weston.
\newblock A unified architecture for natural language processing: Deep neural
  networks with multitask learning.
\newblock In {\em Proceedings of the 25th international conference on Machine
  learning}, pages 160--167. ACM, 2008.

\bibitem{Habibi2017}
Maryam Habibi, Leon Weber, Mariana Neves, David~Luis Wiegandt, and Ulf Leser.
\newblock Deep learning with word embeddings improves biomedical named entity
  recognition.
\newblock {\em Bioinformatics}, 33(14):i37--i48, 2017.

\bibitem{he2017}
Luheng He, Kenton Lee, Mike Lewis, and Luke Zettlemoyer.
\newblock Deep semantic role labeling: What works and what’s next.
\newblock In {\em Proceedings of the 55th Annual Meeting of the Association for
  Computational Linguistics (Volume 1: Long Papers)}, volume~1, pages 473--483,
  2017.

\bibitem{kedzia2013}
Pawe{\l} K{\k{e}}dzia, Maciej Piasecki, Marek Maziarz, and Micha{\l}
  Marci{\'n}czuk.
\newblock Recognising compositionality of multi-word expressions in the wordnet
  oriented perspective.
\newblock In {\em Mexican International Conference on Artificial Intelligence},
  pages 240--251. Springer, 2013.

\bibitem{Kobylinski2018}
{\L}ukasz Kobyliński and Micha{\'l} Wasiluk.
\newblock {Deep Learning in Event Detection in Polish}.
\newblock In {\em Text, Speech and Dialogue (during the review).}, 2018.

\bibitem{Kocon2017}
Jan Koco{\'n} and Micha{\l} Marci{\'n}czuk.
\newblock {Improved Recognition and Normalisation of Polish Temporal
  Expressions}.
\newblock In {\em Proceedings of the International Conference Recent Advances
  in Natural Language Processing, RANLP 2017}, pages 387--393, 2017.

\bibitem{Kocon2015a}
Jan Kocoń and Micha{\l} Marcińczuk.
\newblock {Recognition of Polish Temporal Expressions}.
\newblock {\em Proceedings of the Recent Advances in Natural Language
  Processing}, pages 282--290, 2015.
\newblock Recent Advances in Natural Language Processing (RANLP 2015).

\bibitem{Kocon2016a}
Jan Kocoń and Micha{\l} Marcińczuk.
\newblock {Supervised approach to recognise Polish temporal expressions and
  rule-based interpretation of timexes}.
\newblock {\em Natural Language Engineering}, 23(3):385–418, 2017.

\bibitem{Kocon2015}
Jan Kocoń, Micha{\l} Marcińczuk, Marcin Oleksy, Tomasz Bernaś, and Micha{\l}
  Wolski.
\newblock {T}emporal {E}xpressions in {P}olish {C}orpus {KPW}r.
\newblock {\em Cognitive Studies --- Etudes Cognitives}, 15, 2015.

\bibitem{kocon2014}
Jan Kocoń and Maciej Piasecki.
\newblock {Named Entity Matching Method Based on the Context-Free Morphological
  Generator}.
\newblock In Adam Przepiórkowski and Maciej Ogrodniczuk, editors, {\em
  Advances in Natural Language Processing}, volume 8686 of {\em Lecture Notes
  in Computer Science}, pages 34--44. Springer International Publishing, 2014.

\bibitem{kutuzov2017}
Andrei Kutuzov, Murhaf Fares, Stephan Oepen, and Erik Velldal.
\newblock Word vectors, reuse, and replicability: Towards a community
  repository of large-text resources.
\newblock In {\em Proceedings of the 58th Conference on Simulation and
  Modelling}, pages 271--276. Link{\"o}ping University Electronic Press, 2017.

\bibitem{Lafferty2001}
John~D. Lafferty, Andrew McCallum, and Fernando C.~N. Pereira.
\newblock {Conditional Random Fields: Probabilistic Models for Segmenting and
  Labeling Sequence Data}.
\newblock In {\em Proceedings of the Eighteenth International Conference on
  Machine Learning}, ICML '01, pages 282--289, San Francisco, CA, USA, 2001.
  Morgan Kaufmann Publishers Inc.

\bibitem{Llorens2010a}
Hector Llorens, Estela Saquete, and Borja Navarro-Colorado.
\newblock {Time{ML} Events Recognition and Classification: Learning {CRF}
  Models with Semantic Roles}.
\newblock In {\em Proceedings of the 23\textsuperscript{rd} International
  Conference on Computational Linguistics}, COLING '10, pages 725--733,
  Stroudsburg, PA, USA, 2010. Association for Computational Linguistics.

\bibitem{Marcinczuk2018}
Michał Marcińczuk, Jan Kocoń, and Michał Gawor.
\newblock {Recognition of Named Entities for Polish-Comparison of Deep Learning
  and Conditional Random Fields Approaches}.
\newblock In Maciej Ogrodniczuk and Łukasz Kobyliński, editors, {\em
  Proceedings of the PolEval 2018 Workshop}, pages 77--92. Institute of
  Computer Science, Polish Academy of Science, 2018.

\bibitem{maziarz2016}
Marek Maziarz, Maciej Piasecki, Ewa Rudnicka, Stan Szpakowicz, and Pawe{\l}
  K{\k{e}}dzia.
\newblock plwordnet 3.0--a comprehensive lexical-semantic resource.
\newblock In {\em Proceedings of COLING 2016, the 26th International Conference
  on Computational Linguistics: Technical Papers}, pages 2259--2268, 2016.

\bibitem{mikolov2013a}
Tomas Mikolov, Kai Chen, Greg Corrado, and Jeffrey Dean.
\newblock Efficient estimation of word representations in vector space.
\newblock {\em arXiv preprint arXiv:1301.3781}, 2013.

\bibitem{mikolov2013}
Tomas Mikolov, Ilya Sutskever, Kai Chen, Greg~S Corrado, and Jeff Dean.
\newblock Distributed representations of words and phrases and their
  compositionality.
\newblock In {\em Advances in neural information processing systems}, pages
  3111--3119, 2013.

\bibitem{Mykowiecka2017}
Agnieszka Mykowiecka.
\newblock {Wykorzystanie anotowanego ręcznie korpusu do opracowania metod
  identyfikacji wyrażeń temporalnych i relacji między nimi w tekście.}
\newblock Technical Report CLARIN-PL M12 Z6.2, Instytut Podstaw Informatyki
  PAN, Warszawa, 2017.

\bibitem{Mykowiecka2018}
Agnieszka Mykowiecka.
\newblock {Analiza skuteczności opracowanych metod identyfikacji wyrażeń
  temporalnych.}
\newblock Technical Report CLARIN-PL M18 Z6.3, Instytut Podstaw Informatyki
  PAN, Warszawa, 2018.

\bibitem{mykowiecka2017a}
Agnieszka Mykowiecka, Ma{\l}gorzata Marciniak, and Piotr Rychlik.
\newblock Testing word embeddings for polish.
\newblock {\em Cognitive Studies| {\'E}tudes cognitives}, (17), 2017.

\bibitem{ogrodniczuk2018}
Maciej Ogrodniczuk and Łukasz Kobyliński, editors.
\newblock {\em {Proceedings of the PolEval 2018 Workshop}}, Warsaw, Poland,
  2018. Institute of Computer Science, Polish Academy of Sciences.

\bibitem{piasecki2018}
Maciej Piasecki, Gabriela Czachor, Arkadiusz Janz, Dominik Kaszewski, and
  Pawe{\l} Kedzia.
\newblock Wordnet-based evaluation of large distributional models for polish.
\newblock In {\em Proceedings of the 9th Global Wordnet Conference (GWC 2018).
  Global WordNet Association}, 2018.

\bibitem{pomikalek2011}
Jan Pomik{\'a}lek.
\newblock {\em Removing boilerplate and duplicate content from web corpora}.
\newblock PhD thesis, Masarykova univerzita, Fakulta informatiky, 2011.

\bibitem{przepiorkowski2012}
Adam Przepi{\'o}rkowski.
\newblock {\em Narodowy korpus j{\k{e}}zyka polskiego}.
\newblock Naukowe PWN, 2012.

\bibitem{przepiorkowski2004}
Adam Przepiórkowski.
\newblock {\em The {IPI PAN} Corpus: Preliminary version}.
\newblock 2004.

\bibitem{radziszewski2013}
Adam Radziszewski.
\newblock A tiered crf tagger for polish.
\newblock In {\em Intelligent tools for building a scientific information
  platform}, pages 215--230. Springer, 2013.

\bibitem{radziszewski2011}
Adam Radziszewski and Tomasz Sniatowski.
\newblock Maca-a configurable tool to integrate polish morphological data.
\newblock 2011.

\bibitem{rogalski2016}
Marek Rogalski and Piotr~S Szczepaniak.
\newblock Word embeddings for the polish language.
\newblock In {\em International Conference on Artificial Intelligence and Soft
  Computing}, pages 126--135. Springer, 2016.

\bibitem{santos2015}
Cicero Nogueira~dos Santos and Victor Guimaraes.
\newblock Boosting named entity recognition with neural character embeddings.
\newblock {\em arXiv preprint arXiv:1505.05008}, 2015.

\bibitem{Sauri2006}
Roser Saur{\'{\i}}, Jessica Littman, Robert Gaizauskas, Andrea Setzer, and
  James Pustejovsky.
\newblock {TimeML Annotation Guidelines, Version 1.2.1}, 2006.

\bibitem{Strotgen2013}
Jannik Str{\"{o}}tgen and Michael Gertz.
\newblock {Multilingual and cross-domain temporal tagging}.
\newblock {\em Language Resources and Evaluation}, 47(2):269--298, 2013.

\bibitem{Uzzaman2013}
Naushad UzZaman, Hector Llorens, Leon Derczynski, James Allen, Marc Verhagen,
  and James Pustejovsky.
\newblock {Semeval-2013 task 1: Tempeval-3: Evaluating time expressions,
  events, and temporal relations}.
\newblock In {\em Second Joint Conference on Lexical and Computational
  Semantics (* SEM), Volume 2: Proceedings of the Seventh International
  Workshop on Semantic Evaluation (SemEval 2013)}, volume~2, pages 1--9, 2013.

\bibitem{Wendelberger2015}
Michał Wendelberger.
\newblock Automatic extraction and classification of collocations from polish
  corpora.
\newblock Master's thesis, Wrocław University of Science and Technology, 2015.

\bibitem{wolinski2014}
Marcin Woli{\'n}ski.
\newblock Morfeusz reloaded.
\newblock In {\em Proceedings of the Ninth International Conference on Language
  Resources and Evaluation, LREC}, pages 1106--1111, 2014.

\bibitem{young2018}
Tom Young, Devamanyu Hazarika, Soujanya Poria, and Erik Cambria.
\newblock Recent trends in deep learning based natural language processing.
\newblock {\em ieee Computational intelligenCe magazine}, 13(3):55--75, 2018.

\bibitem{yu2014}
Lei Yu, Karl~Moritz Hermann, Phil Blunsom, and Stephen Pulman.
\newblock Deep learning for answer sentence selection.
\newblock {\em arXiv preprint arXiv:1412.1632}, 2014.

\end{thebibliography}

\end{document}